\documentclass[letterpaper, 10 pt, journal, twoside]{IEEEtran}

\IEEEoverridecommandlockouts                              

\renewcommand{\baselinestretch}{0.99}

\usepackage{graphicx}
\usepackage{comment}
\usepackage{amsmath,amssymb,bm} 
\usepackage{color}
\usepackage{cprotect}
\usepackage[noend]{algorithmic}
\usepackage[]{algorithm}
\usepackage{xcolor}
\usepackage{multirow}

\title{
Learning Occupancy Priors of Human Motion \\ from Semantic Maps of Urban Environments
}

\author{Andrey Rudenko$^{1,3}$, Luigi Palmieri$^{1}$, Johannes Doellinger$^{2}$, Achim J. Lilienthal$^{3}$ and Kai O. Arras$^{1}$%
	\thanks{Manuscript received: July 11, 2020; Revised: November 4, 2020; Accepted February 4, 2021.}
	\thanks{This paper was recommended for publication by Editor Tamim Asfour upon evaluation of the Associate Editor and Reviewers' comments. This work has been partly funded from the European Union's Horizon 2020 research and innovation programme under grant agreement No 732737 (ILIAD).}
	\thanks{$^{1}$A. Rudenko, L. Palmieri and K.O. Arras are with Bosch Corporate Research, Renningen, Germany {\tt\footnotesize \{andrey.rudenko, luigi.palmieri, kaioliver.arras\}@de.bosch.com}}%
	\thanks{$^{2}$J. Doellinger is with the Bosch Center for Artificial Intelligence, Renningen, Germany {\tt\footnotesize johannes.doellinger@de.bosch.com}}%
	\thanks{$^{3}$A. Rudenko and A.J. Lilienthal are with the Mobile Robotics and Olfaction Lab, \"Orebro University, Sweden {\tt\footnotesize Achim.Lilienthal@oru.se}}
	\thanks{Digital Object Identifier (DOI): see top of this page.}
}

\begin{document}

\maketitle

\markboth{IEEE Robotics and Automation Letters. Preprint Version. Accepted February 2021}
{Rudenko \MakeLowercase{\textit{et al.}}: Learning Occupancy Priors of Human Motion from Semantic Maps of Urban Environments} 

\begin{abstract}
	
Understanding and anticipating human activity is an important capability for intelligent systems in mobile robotics, autonomous driving, and video surveillance. While learning from demonstrations with on-site collected trajectory data is a powerful approach to discover recurrent motion patterns, generalization to new environments, where sufficient motion data are not readily available, remains a challenge. In many cases, however, semantic information about the environment is a highly informative cue for the prediction of pedestrian motion or the estimation of collision risks. In this work, we infer occupancy priors of human motion using only semantic environment information as input. To this end we apply and discuss a traditional Inverse Optimal Control approach, and propose a novel one based on Convolutional Neural Networks (CNN) to predict future occupancy maps. Our CNN method produces flexible context-aware occupancy estimations for semantically uniform map regions and generalizes well already with small amounts of training data. Evaluated on synthetic and real-world data, it shows superior results compared to several baselines, marking a qualitative step-up in semantic environment assessment.

\end{abstract}


\section{Introduction}
\label{sec:introduction}

\IEEEPARstart{U}{nderstanding} and predicting human motion is an increasingly popular subject of research with the goal of improving the safety and efficiency of autonomous systems in spaces shared with people. 
Application areas include mobile service robots, intelligent vehicles, collaborative production assistants, video surveillance, or urban city planning. 
Human motion in these scenarios is influenced by many factors, including other agents in the scene and the environment itself, which can be represented by a topometric map and semantic information \cite{rudenko2020surveyIJRR}. 
While indoor human navigation is often motivated by avoiding collisions with static and dynamic obstacles, surface semantics have a strong impact in outdoor (e.g. urban) environments.
For instance, pedestrians walk most of the time on sidewalks, sometimes on streets, unpaved areas and greenspaces, and very rarely over obstacles. 
Modeling the influence of semantics is a challenging task, typically approached with data-driven methods using human trajectory data in a given environment \cite{ellis2009modelling,wang2015modeling,kucner2017enabling} without knowing the goal of the target agent. 
Powerful in scenes known beforehand, such approaches may suffer from poor generalization to never-seen or changing environments where no data is available.

\begin{figure}[t]
	\begin{center}
		\vspace{5pt}
		\includegraphics[angle=270,width=0.45\columnwidth]{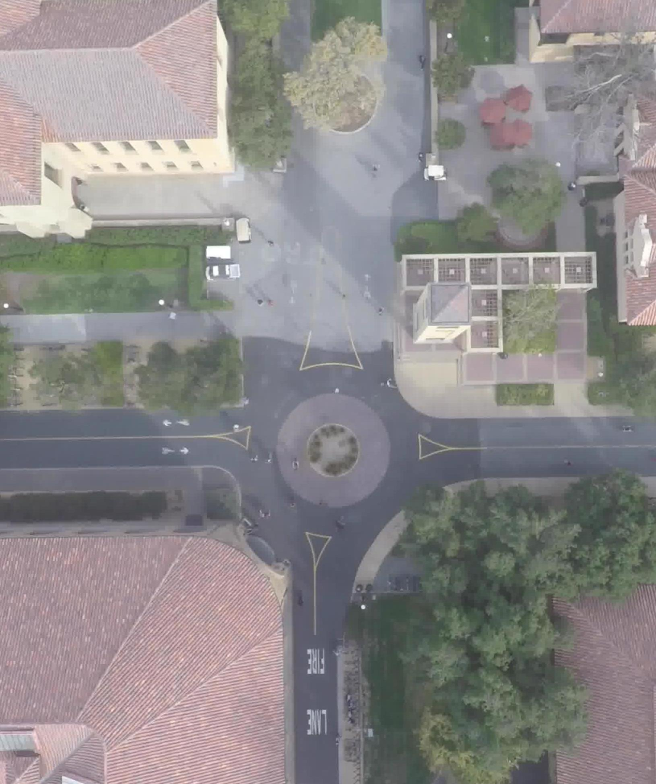} 
		\includegraphics[angle=270,width=0.45\columnwidth]{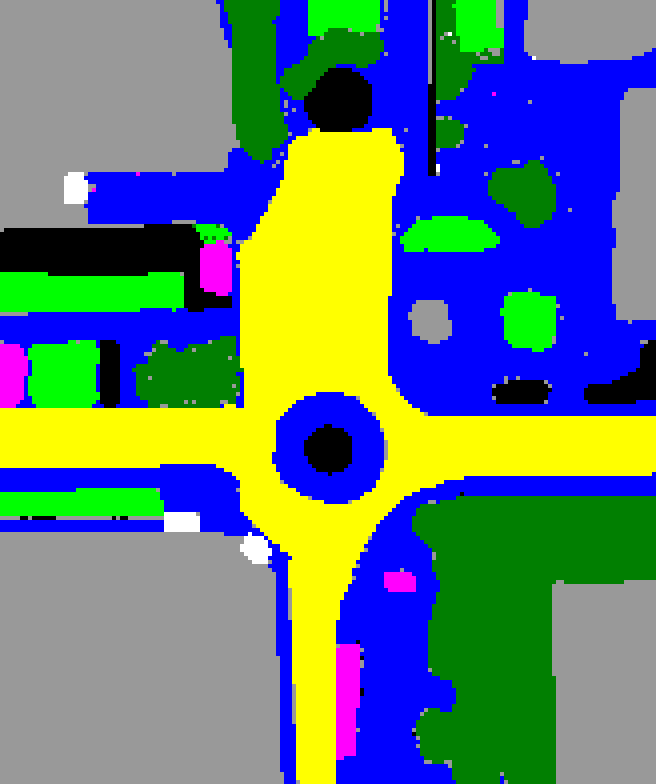} \\
		\vspace{3pt}
		\includegraphics[angle=270,width=0.45\columnwidth]{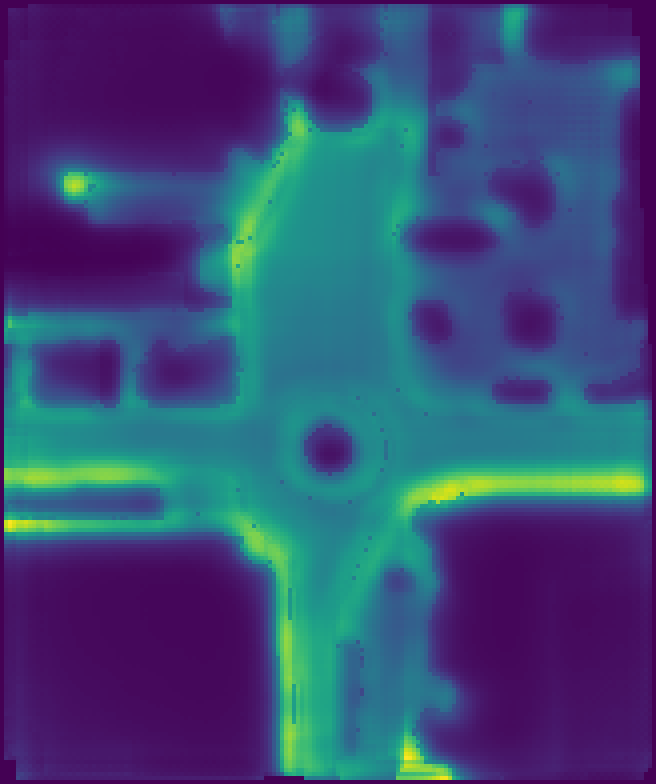}
		\includegraphics[angle=270,width=0.45\columnwidth]{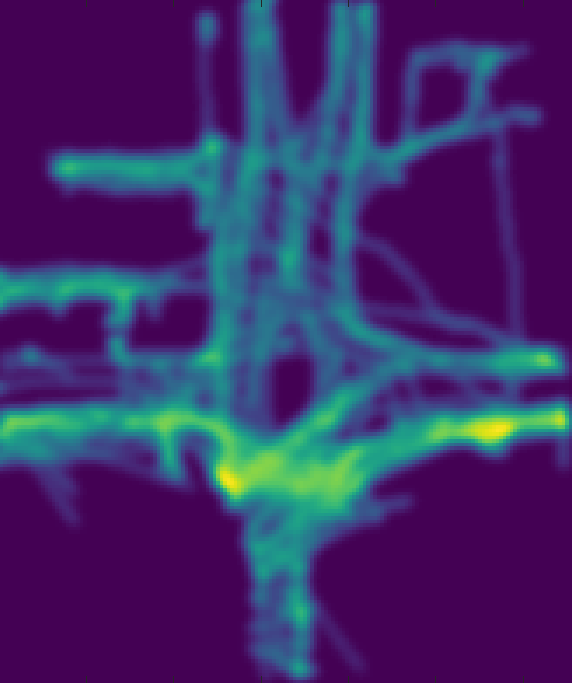}
	\end{center}
	\vspace{-5pt}
	\caption{Predicting occupancy priors in semantically-rich urban environments. {\bf Top left:} an urban scene from the Stanford Drone dataset. {\bf Top right:} semantic map of the environment. {\bf Bottom left:} CNN-predicted occupancy distribution priors of walking people in the environment, encoded with a heatmap: warmer colors correspond to states with higher probability of observing pedestrians. {\bf Bottom right:} ground truth occupancy distribution.}
	\label{fig:cover}
	\vspace{-8pt}
\end{figure}

In this paper, we research the possibility of inferring occupancy priors of walking people in previously unseen places with limited input, namely using only the semantic map of the area. A prior occupancy distribution is intuitively interpretable and beneficial for a large variety of applications, such as improved goal estimation \cite{rehder2015goal,wu2018probabilistic} (for instance in Fig.~\ref{fig:cover}, not all walking directions are likely to be the goal of a person) and crossing intention recognition in autonomous driving tasks \cite{pool2019context,huIVS2018}, where possible ``illegal crosswalks'' could be easily detected. The usage of semantic map-based occupancy priors may further improve the accuracy of map-based motion prediction approaches \cite{karasev2016intent}, which often assume constant priors for each semantic class. Such occupancy estimation can guide a cleaning robot towards more heavily used areas, or a service robot in search of people to assist.

Traditionally, Inverse Reinforcement Learning (IRL) has been used to learn semantic preferences of walking people in urban and semantically-rich environments \cite{kitani2012activity,rhinehart2018first,vanderHeiden2019safecritic}. It is indeed possible to use the learned preferences to simulate trajectories and infer the prior occupancy distribution in a new environment. In this work we review the IRL methodology, applied to the occupancy prior distribution inference, and discuss its limitations.
In order to address them, we introduce a novel extension to the recent method by Doellinger et al.~\cite{doellinger2018predicting}, which uses a Convolutional Neural Network (CNN) to predict average occupancy maps indoors, with semantic map input for the urban scenes. 
We train our method on scenes from the Stanford Drone Dataset \cite{robicquet2016learning}, as well as on simulated environments. 
In comparison to several baselines, our CNN method predicts much more accurate prior occupancy priors in terms of KL-divergence to the ground truth distributions. Crucially, it makes a qualitative improvement of estimating flexible occupancy priors for semantically-uniform areas by considering local context and interconnections between different semantic regions.

In summary, we make the following contributions:
\begin{itemize}
	\item We analyze and discuss state-of-the-art methodology for inferring occupancy prior distribution in semantically-rich urban environments.
	\item Addressing the limitations of the prior art, we propose a novel method based on Convolutional Neural Networks (displayed in Fig.~\ref{fig:cover} and \ref{fig:network}).
	\item We execute a thorough comparison of the discussed methods with several baselines, and show qualitative and quantitative improvement of the KL-divergence scores when using our CNN method.
\end{itemize}


The paper is structured as follows: in Sec.~\ref{sec:related_work} we review the related work, in Sec.~\ref{sec:approach} we detail the proposed solutions and in Sec.~\ref{sec:experiments} we describe the training and evaluation. Results are presented in Sec.~\ref{sec:results}, and a discussion in Sec.~\ref{sec:discussion} concludes the paper.

\section{Related Work}
\label{sec:related_work}
The ability to understand a human environment and its affordances is useful in a number of tasks where intelligent autonomous systems need to reason on observed events, anticipate future events, evaluate risks and act in a dynamic world. 
Examples include person and group tracking \cite{pellegrini2009you,lau2009tracking,linder2016multi}, in particular over a camera network with non-overlapping fields of view, human-aware motion planning \cite{Foka2010,Bai2015,palmieri2017kinodynamic}, motion behavior learning \cite{okal2016learning}, human motion prediction \cite{chung2012incremental,Rudenko2018iros}, human-robot interaction \cite{lasota2017survey}, video surveillance \cite{alahi2016social} or collision risk assessment \cite{lo2019robust}. 
Apart from basic geometric properties of the workspace, its semantics have a large impact on human motion in these tasks.
Modeling this impact is challenging, therefore a popular approach is learning the motion patterns directly from data without explicitly specifying semantic features \cite{ellis2009modelling,trautman2010unfreezing,wang2015modeling,kucner2017enabling}. 
However, many of those methods either need additional training input in new environments or experience transfer issues. To the best of our knowledge, only a few methods explicitly highlight the performance in new environments outside the training scenario \cite{kitani2012activity,ballan2016knowledge,srikanth2019infer,shenTransferable2018}.
As we will later describe, our approach explicitly uses only semantic maps as input and there will be no need to adapt the learned models to new environments, described with the same semantic classes.

Modeling the effect of surface classes on human motion was mainly used in reactive approaches such as \cite{coscia2018long} and planning-based approaches \cite{kitani2012activity,karasev2016intent}. These methods leverage semantic segmentation tools for understanding and detecting the semantics in the environment. Several approaches exist to segment available semantics \cite{munoz2010stacked,badrinarayanan2017segnet,wu2019wider} and to build semantic maps of the environment \cite{gan2019bayesian,grimmett2015integrating} -- a prerequisite to the methods presented in this paper.
We build on those methods to predict areas frequently used by pedestrians based on the semantic class of the surface.  


Several Inverse Reinforcement Learning (IRL, or Inverse Optimal Control, IOC) approaches make use of semantic maps for predicting future human motion \cite{karasev2016intent,kitani2012activity,shenTransferable2018}. In particular, they use the semantic maps for encoding the features of the reward function. 
However, these IRL approaches are limited to one weight per feature and thus do not generalize well to new environments or heteregenous datasets with different geometries \cite{yu2019metainverse,xu2018learning}. 
In this work, we review an adaption of the Maximum Entropy IRL algorithm \cite{ziebart2008maximum} to the task of occupancy priors estimation, and compare our CNN-based approach to it.


\section{Occupancy Priors Estimation in Urban Environments}
\label{sec:approach}

In this work, we study the problem of estimating occupancy priors of walking humans in semantically-rich urban environments. 
The problem is formulated as follows: given a grid-map of the environment $\mathcal{M}$ with associated feature responses $\bm{f}(s)=[f_1(s),\dots, f_K(s)],~\sum_{k=1}^{K} f_k(s) = 1$ for each state $s \in \mathcal{M}$ over the set of $K$ semantic classes, we seek to estimate the probability $p(s)$ of a walking human being observed in this state.

If we assume having access to a large set of trajectories $\bm{\mathcal{T}}_\mathcal{M}$ in $\mathcal{M}$, the problem of estimating $p(s)$ can be solved by counting visitation frequencies in each state:

\begin{equation}
	p(s) = \frac{D(s)}{\sum_{s' \in \mathcal{M}} D(s')} \Bigr\vert_{\bm{\mathcal{T}}_\mathcal{M}},
	\label{eq:occupancy_estimation}
\end{equation}
where $D(s)$ is visitation count of state $s$ over all trajectories $\bm{\mathcal{T}}_\mathcal{M}$. In this paper, we estimate this distribution in environments where no trajectory data are available. One natural way to overcome the lack of trajectories is to simulate them, in particular using learned human walking preferences. To this end, in Sec.~\ref{sec:ioc} we first review an Inverse Optimal Control (IOC) method \cite{kitani2012activity} for predicting motion trajectories in semantic environments, and discuss its applicability and limitations. Then, in Sec.~\ref{sec:cnn}, we propose a novel approach based on Convolutional Neural Networks, which is an extension of the occupancy prior estimation method by Doellinger et al. \cite{doellinger2018predicting}. For this task, we assume a semantic map of the environment $\bm{f}(\mathcal{M})$, or a method to extract it, to be available. Without loss of generality and for the sake of visual clarity, the states in this paper are represented with one-hot vectors, i.e. $\forall s~\exists~k~\text{s.t.}~f_k(s)=1$ and $\forall j \neq k: f_j(s)=0$.

\subsection{Inverse Optimal Control on Multiple Maps (IOCMM)}
\label{sec:ioc}

Both Reinforcement Learning (RL) and Inverse RL or Inverse Optimal Control (IOC) frameworks deal with modeling optimal behavior of an agent, operating in a stochastic world $\mathcal{S}$ and collecting rewards $\mathcal{R}$ on the way to their goal state $s_g \in \mathcal{S}$. An agent's behavior is encoded in a policy $\pi(a|s)$, which maps the state $s \in \mathcal{S}$ to a distribution over actions $a \in \mathcal{A}$. 
When the reward function is not known beforehand, which is the case in many real-world applications, one possiblity is  to learn it from a set of observations $\bm{\mathcal{T}}$ with an IOC method. In this case, the reward function is parametrized by a set of parameters $\bm{\theta}$.

Modeling the behavior of an agent navigating in the environment, which is described with a set of features $\bm{f}(s)$ for each state, suits the problem of recovering occupancy priors from semantic map inputs well. Prior art, however, has not dealt with abstract quantities, such as occupancy expectations, focusing rather on the policy of an individual agent \cite{ziebart2009planning} or multiple agents jointly \cite{kuderer2012feature}. In our work we adapt the IOC framework to this task. As our IOC implementation is based on \cite{kitani2012activity}, we give a short summary of their approach in this section.

MDP-based Maximum Entropy Inverse Reinforcement Learning (MaxEnt IRL) \cite{ziebart2008maximum} assumes that the observed motion of agents is generated by a stochastic motion policy, and seeks to estimate this policy with maximum likelihood to the available demonstrations. The reward an agent gets in state $s$ is linear with respect to the feature responses in that state: $\mathcal{R}(s,{\bf \bm{\theta}}) = r_0 + \bm{\theta}^T \bm{f}(s)$, where $r_0>0$ is the base reward of a transition and $\bm{\theta}$ is a set of weights or \emph{costs} of the semantic classes: $\sum_{k=1}^{K} \theta_k = 1, \theta_k \in [0,1]$. Given $\mathcal{R}$, the distribution over the sequence of states $\bm{s}$ is defined as

\begin{equation}
	p(\bm{s},\bm{\theta}) = \frac{\prod_t e^{\mathcal{R}(s_t, \bm{\theta})}}{Z(\bm{\theta})} = \frac{e^{\sum_t r_0 + \bm{\theta}^T \bm{f}(s_t)}}{Z(\bm{\theta})}.
	\label{eq:maxent_dist_over_trajectories}
\end{equation}

\begin{algorithm}[b]
	\centering
	\caption{Backward pass}\label{alg:backward_pass}
	\footnotesize
	\begin{algorithmic}[1]
		\STATE {\bf function} BackwardPass($\mathcal{T}^i_\mathcal{M},\bm{\theta}$)
		\STATE{$V(s) \gets -\infty$}
		\FOR{$n = N,\dots,1$}
		\STATE{$V^{(n)}(s_g) \gets 0$}
		\STATE{$Q^{(n)}(s,a) \gets \mathcal{R}(s,\bm{\theta}) + E_{P^{s'}_{s,a}}[V^{(n)}(s')]$}
		\STATE{$V^{(n-1)}(s) \gets $ {\normalfont softmax}$_a~Q^{(n)}(s,a)$}
		\ENDFOR
		\STATE $\pi(a|s) \gets e^{\alpha(Q(s,a)-V(s))}$
		\RETURN $\pi$
	\end{algorithmic}
\end{algorithm}

\begin{algorithm}[b]
	\centering
	\caption{Forward pass}\label{alg:forward_pass}
	\footnotesize
	\begin{algorithmic}[1]
		\STATE {\bf function} ForwardPass($\mathcal{T}^i_\mathcal{M},\pi$)
		\STATE{$D \gets 0$}
		\FOR{$n = 1,\dots,N$}
		\STATE{$s \gets s_0$}
		\WHILE{$s \neq s_g$}
		\STATE $D(s) \gets D(s)+1$
		\STATE $s' \gets \pi(a|s)$
		\STATE $s \gets s'$
		\ENDWHILE
		\ENDFOR
		\STATE  $\hat{\text{\bfseries f}}_\theta \gets \sum_s \bm{f}(s)D(s)$
		\RETURN $\hat{\text{\bfseries f}}_\theta$
	\end{algorithmic}
\end{algorithm}

Finding the optimal $\bm{\theta}^*$ vector is equivalent to maximizing the entropy of $p(\bm{s},\bm{\theta})$ in Eq.~\ref{eq:maxent_dist_over_trajectories} while matching the semantic class feature counts of the training trajectories. An iterative procedure based on the exponentiated gradient descent of the log-likelihood $\mathcal{L} \triangleq \log p(\bm{s}|\bm{\theta})$ is described by Kitani et al. in \cite{kitani2012activity}. The gradient $\nabla \mathcal{L}_\theta$ is computed as the difference between the \emph{empirical} mean feature count $\bar{\text{\bfseries f}} = \frac{1}{|\bm{\mathcal{T}}|} \sum_{i}^{|\bm{\mathcal{T}}|} \bm{f}(\mathcal{T}^i)$, i.e. the average features accumulated over the $\bm{\mathcal{T}}$ training trajectories in the map $\mathcal{M}$, and the \emph{expected} mean feature count $\hat{\text{\bfseries f}}_\theta$, the average features accumulated by trajectories generated by the current parameters $\bm{\theta}$: $\nabla \mathcal{L}_\theta = \bar{\text{\bfseries f}} - \hat{\text{\bfseries f}}_\theta$. The weight vector is then updated as

\begin{equation}
	\bm{\theta} \gets \bm{\theta} e^{\lambda \nabla \mathcal{L}_\theta},
	\label{eq:theta_update}
\end{equation}
where $\lambda$ is the learning rate, and the expected mean feature count $\hat{\text{\bfseries f}}_\theta$ is computed using an iterative algorithm described below.

\begin{algorithm}[b]
	\centering
	\caption{IOCMM}\label{alg:iocmm}
	\footnotesize
	\begin{algorithmic}[1]
		\STATE $\bm{\theta} \gets 1/K$
		\REPEAT
		\STATE $\hat{\text{\bfseries f}}_\theta \gets 0,~\bar{\text{\bfseries f}} \gets 0$
		\STATE {\normalfont Batch} $B_m$ {\normalfont maps}
		\FOR{$m = 1,\dots,B_m$}
		\STATE $\bm{\mathcal{T}} \gets${\normalfont ~Batch} $B_t$ {\normalfont trajectories from} $\mathcal{M}_m$
		\STATE $\bar{\text{\bfseries f}} \gets \bar{\text{\bfseries f}} + \frac{1}{|\bm{\mathcal{T}}|} \sum_{i}^{|\bm{\mathcal{T}}|} \bm{f}(\mathcal{T}^i)$
		\STATE $\mathcal{R}(s,{\bf \bm{\theta}}) \gets r_0 + \bm{\theta}^T \bm{f}(s)$
		\FOR{$i = 1,\dots,B_t$}
		\STATE $\pi \gets$\verb!BackwardPass!$(s_g)$
		\STATE $\hat{\text{\bfseries f}}_\theta \gets \hat{\text{\bfseries f}}_\theta + $ \verb!ForwardPass!$(s_0,\pi)$
		\ENDFOR
		\ENDFOR
		\STATE $\bar{\text{\bfseries f}} \gets \verb!normalize!(\bar{\text{\bfseries f}})$
		\STATE $\hat{\text{\bfseries f}}_\theta \gets \verb!normalize!(\hat{\text{\bfseries f}}_\theta)$
		\STATE $\nabla \mathcal{L}_\theta \gets  \bar{\text{\bfseries f}} - \hat{\text{\bfseries f}}_\theta$
		\STATE{$\bm{\theta} \gets \bm{\theta} e^{\lambda \nabla \mathcal{L}_\theta}$}
		\UNTIL{$||\nabla \mathcal{L}_\theta||<\epsilon$}
	\end{algorithmic}
\end{algorithm}

The algorithm iterates backward and forward passes, detailed in Alg.~\ref{alg:backward_pass} and \ref{alg:forward_pass} respectively. The \emph{backward pass} uses the current $\bm{\theta}$ vector to compute the value function $V(s)$ for each state $s$ in $\mathcal{M}$ given the goal state $s_g$ -- the final state of the trajectory $\mathcal{T}^i_\mathcal{M} \in \bm{\mathcal{T}}_\mathcal{M}$. A stochastic motion policy $\pi_\theta (a|s)$ to reach $s_g$ in $\mathcal{M}$ under $\mathcal{R}(s,\bm{\theta})$ is then computed and used in the \emph{forward pass} to simulate several trajectories from $s_0$ to $s_g$, where $s_0$ is the initial state of $\mathcal{T}^i_\mathcal{M}$. The expected mean feature count is computed as a weighted sum of feature counts $\hat{\text{\bfseries f}}_\theta = \sum_s \bm{f}(s)D(s)$ in the simulated trajectories, and the backward-forward iteration is repeated for a batch of trajectories in $\bm{\mathcal{T}}_\mathcal{M}$. The $\bm{\theta}$ vector is updated as in Eq.~\ref{eq:theta_update} using the cumulative $\hat{\text{\bfseries f}}_\theta$ for the trajectories in the batch, and the algorithm is iterated until the gradient $\nabla \mathcal{L}_\theta$ reaches zero.
In order to learn from multiple maps, including those where only a subset of $K$ features is present, we run the backward and forward passes for a batch of trajectories \emph{in a batch of maps}, accumulating the visitation counts $D(s)$ across several maps. The resulting Inverse Optimal Control on Multiple Maps (IOCMM) method is detailed in Alg.~\ref{alg:iocmm}.

Having obtained the optimal $\bm{\theta}^*$ weights, it is possible to compute the reward $\mathcal{R}(s,{\bf \bm{\theta}^*})$ and simulate trajectories in any environment which is described by a subset of $K$ semantic features. By simulating semantic-aware trajectories, an average visitation count for each state, normalized across all states in $\mathcal{M}$, yields the occupancy probability $p(s)$, as in Eq.~\ref{eq:occupancy_estimation}. Apart from the $\bm{\theta}^*$ vector, this simulation depends on the distributions from which the initial and goal states $s_{0}$ and $s_{g}$ (hereinafter denoted $s_{0,g}$) are drawn: since the algorithm is inherently unaware of the semantics behind classes, omitting this step may result in $s_{0,g}$ generation inside of obstacles or other high-cost areas. To counteract this issue, we consider two strategies: (1) directly learn probabilities to sample the start or goal position in a state $s$, conditioned on the semantic class $\bm{f}(s)$ of the state: $p(s_{0,g}|\bm{f}(s))$, and (2) generate the $s_{0,g}$ only from low-cost regions with the softmax function over the estimated cost of the state: $p(s_{0,g}) \sim e^{-\mathcal{R}(s,\bm{\theta}^*)/\tau}$. Furthermore, to generate long trajectories spanning across the map, both distributions (1) and (2) are scaled linearly proportional to the distance between the $s_{0,g}$ and the center of the map.

\subsubsection{Analysis and Discussion}
\label{sec:ioc:discussion}


While delivering adequate results in our experiments, as we show in Sec.~\ref{sec:experiments}, the IOCMM approach to occupancy priors estimation has an inherent drawback. With rigid costs of a semantic class $k$, defined by the corresponding $\theta_k$ weight, the IOCMM method cannot produce flexible estimations for a spatial region given its position in a wider topological structure. For instance, if we assume that the grass surface is walkable, then it will have a low cost and predicted people would largely ignore paved paths in a park. However, this behavior is probably not confirmed in the training data, which will increase the costs of the grass regions, potentially making them not traversable in some cases where such behavior is expected. Controlled by one $\theta_k$ parameter, the cost of the semantic class stays constant over the entire map. Similarly, learning to step on the road surface in places where this behavior is unavoidable will inevitably lead to decreasing the costs of the road surface \emph{everywhere} in the map. Learning such flexible behavior requires reasoning over the local context and interconnections between different semantic attributes and the surface.
To this end, we propose our Convolutional Neural Network-based approach ``Semantic Map-Aware Pedestrian Prediction" (\emph{semapp}), described in the next section.

\begin{figure}[t]
	\begin{center}
		\vspace{5pt}
		\includegraphics[width=0.999\columnwidth]{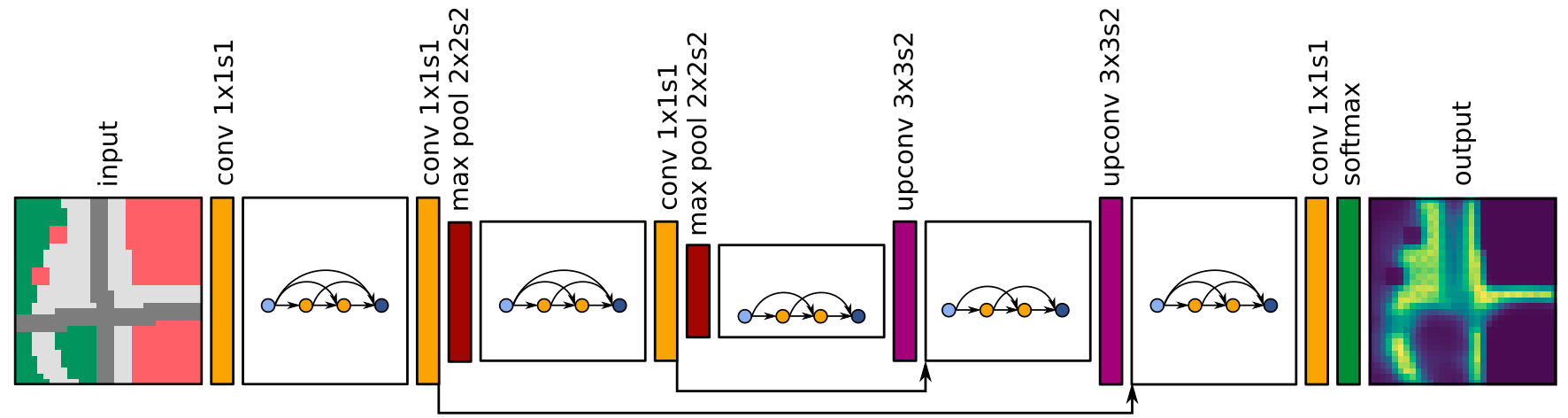}
	\end{center}
	\vspace{-8pt}
	\cprotect\caption{Structure of the \emph{semapp} network. Input semantic tensor of size \verb|<height>x<width>x<features>| 
		is downsampled twice before passing through a bottleneck and getting upsampled to the original resolution again. Black arrows indicate skip connections. The kernel sizes and strides are denoted as \verb|<kernel width>x<kernel height>s<stride>|.}
	\label{fig:network}
	\vspace{-8pt}
\end{figure}

\subsection{Semantic Map-Aware Pedestrian Prediction (semapp)}
\label{sec:cnn}

Convolutional Neural Networks (CNNs) have shown great successes for operations on map data, such as semantic segmentation \cite{munoz2010stacked} or value function estimation for deep reinforcement learning \cite{tamar2016value}. 
For our task of predicting occupancy distributions of walking humans in semantic environments, we need a method to map the feature responses $\bm{f}(\mathcal{M})$ to probabilities $p(s)$. 
To this end, we extend the network to predict occupancy values in semantics-free geometric environments \cite{doellinger2018predicting}. 
This network, based on the FC-DenseNet architecture \cite{jegou2017one}, has reasonably few parameters which helps to avoid overfitting when training on limited amounts of data. Experiments with different architectures have been made in \cite{doellinger2018predicting} but the authors have found their results to be very robust to such changes. We thus decided to perform no further optimization on the network architecture. The method in \cite{doellinger2018predicting} is referred to as Map-Aware Pedestrian Prediction (\emph{mapp}), therefore we call our extension ``Semantic~mapp", or \emph{semapp}.
Extending the architecture from \cite{doellinger2018predicting} to semantic inputs by changing the input from one binary input channel to one channel for each semantic class allows the network to differentiate between pedestrians walking on grass, sidewalks and streets additional to avoiding obstacles. The architecture is outlined in Fig.~\ref{fig:network}.

The network directly outputs the map-sized tensor with the occupancy distribution, so, unlike IOCMM, semapp requires no trajectory simulation for inference. Consequently, for training we convert the trajectories $\bm{\mathcal{T}}_\mathcal{M}$ in each map $\mathcal{M}$ into the occupancy distribution using Eq.~\ref{eq:occupancy_estimation}. This conversion itself is not without meaning: trajectories, as compared to the processed occupancy distribution, contain additional
temporal information. However, this information is not necessarily relevant for the task at hand -
in fact, we are deliberately discarding the temporal aspect of human motion, inferring instead the generalized prior of observing a person in any state of the environment. Using
directly the distribution emphasises the relationship between the topology of the
environment and the desired occupancy priors. Furthermore, it relaxes the requirements to the data itself: detections are sufficient, and there is no need for continuous tracks.

\begin{figure}[t]
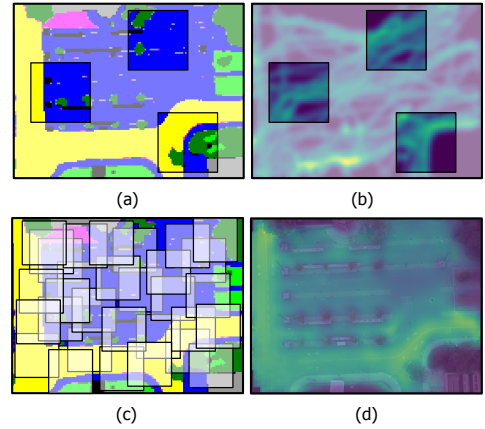

	\begin{center}
		\vspace{7pt}
		\includegraphics[width=0.71\columnwidth]{media/iocmm_larger_maps_1.pdf}\\
		\vspace{3pt}
		\includegraphics[width=0.71\columnwidth]{media/iocmm_larger_maps_2.pdf}
	\end{center}
	\vspace{-12pt}
	\cprotect\caption{{\bf (a)} CNN training on crops from larger maps in the Stanford Drone Dataset, {\bf (b)} corresponding crops from the ground truth distribution. {\bf (c)} Inference for a large map by averaging the inference for random crops in the test image, {\bf (d)} full predicted occupancy distribution from 500 crops.}
	\label{fig:large_maps}
	\vspace{-8pt}
\end{figure}

Since the network operates on map crops of fixed size, we decompose a larger input image into a number of random crops of appropriate size and then rebuilt the final distribution $p(s)$ for state $s$ as an average of predicted occupancy values of $s$ in all crops which include that state (see Fig.~\ref{fig:large_maps}). In this case random crops, as compared to regular grids, remove the aliasing issues from combining adjacent crops.

\section{Experiments}
\label{sec:experiments}

In this section, we give an overview of the training data (Sec.~\ref{sec:experiments:datasets}) and the experiments' design, as well as details on the training and baseline implementation (Sec.~\ref{sec:experiments:training}).

\subsection{Datasets}
\label{sec:experiments:datasets}

We evaluate all methods on two datasets of human trajectories in semantically-rich environments: the Stanford Drone Dataset \cite{robicquet2016learning} and a set of simulated maps. Both datasets are summarized in Table~\ref{tab:datasets}.

\begin{figure}[t]
	\begin{center}
		\vspace{5pt}
		\includegraphics[width=0.149\columnwidth]{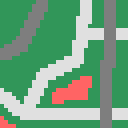}
		\includegraphics[width=0.149\columnwidth]{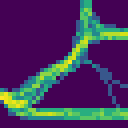}
		\hspace{1pt}
		\includegraphics[width=0.149\columnwidth]{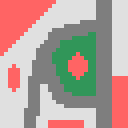}
		\includegraphics[width=0.149\columnwidth]{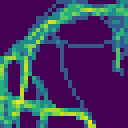}
		\hspace{1pt}
		\includegraphics[width=0.149\columnwidth]{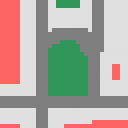}
		\includegraphics[width=0.149\columnwidth]{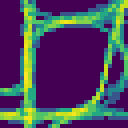} \\
		\vspace{6pt}
		\includegraphics[width=0.149\columnwidth]{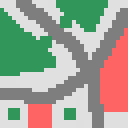}
		\includegraphics[width=0.149\columnwidth]{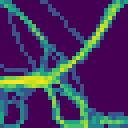}
		\hspace{1pt}
		\includegraphics[width=0.149\columnwidth]{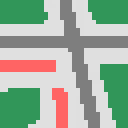}
		\includegraphics[width=0.149\columnwidth]{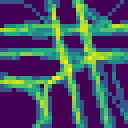}
		\hspace{1pt}
		\includegraphics[width=0.149\columnwidth]{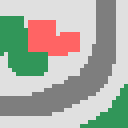}
		\includegraphics[width=0.149\columnwidth]{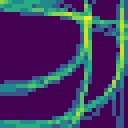} \\
		\vspace{6pt}
		\includegraphics[width=0.149\columnwidth]{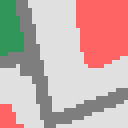}
		\includegraphics[width=0.149\columnwidth]{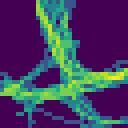}
		\hspace{1pt}
		\includegraphics[width=0.149\columnwidth]{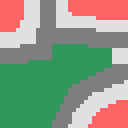}
		\includegraphics[width=0.149\columnwidth]{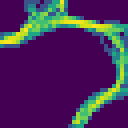}
		\hspace{1pt}
		\includegraphics[width=0.149\columnwidth]{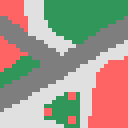}
		\includegraphics[width=0.149\columnwidth]{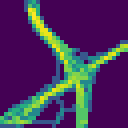}
	\end{center}
	\vspace{-8pt}
	\caption{Training examples from the U4 dataset: each pair shows the semantic map on the left and the ground truth occupancy distribution on the right. Semantic classes include vehicle road in {\bf dark gray}, pedestrian areas and sidewalks in {\bf light gray}, unpaved areas and grass in {\bf green} and obstacles in {\bf red}.}
	\label{fig:simulated_dataset}
	\vspace{-8pt}
\end{figure}

To prove the concept of learning occupancy distributions from semantic maps, we created the ``U4'' dataset which includes 80 hand-crafted maps of simulated \underline{U}rban environments with \underline{four} semantic classes (sidewalk, grass, road and obstacle) and manually marked $\sim$30 trajectories in each map. In this dataset, we pay particular attention to ``illegal crosswalk'' detection, i.e. such scenes where global topology of the environment encourages people to step onto the driveway and cross it. Additionally, as people often tend to cut sharp corners by walking over grass, such behavior is also included in this dataset. Several scenes from U4 are shown in Fig.~\ref{fig:simulated_dataset}.

\setlength{\tabcolsep}{4pt}
\begin{table}
	\vspace{8pt}
	\begin{center}
		\caption{Datasets summary}
		\label{tab:datasets}
		\scriptsize
		\begin{tabular}{lll}
			\hline\noalign{\smallskip}
			Dataset & U4 & Stanford Drone \\
			\noalign{\smallskip}
			\hline
			\noalign{\smallskip} 
			Number of maps & 80 & 25 \\
			Map size in pixels & $32 \times 32$ & $\sim146 \times 152$  \\
			Resolution & simulation & 0.4 m \\
			Number of trajectories per map & $\sim30$ & $\sim132$ \\
			Number of semantic classes in the dataset  & 4 & 9 \\
			\hline
		\end{tabular}
	\end{center}
\end{table}
\setlength{\tabcolsep}{1.4pt}

\setlength{\tabcolsep}{4pt}
\begin{table*}
	\vspace{8pt}
	\begin{center}
		\caption{IOCMM and semapp parameters used for training and inference in the U4 and Stanford Drone datasets}
		\label{tab:hyperparamaters}
		\scriptsize
		\begin{tabular}{lllllll}
			\hline\noalign{\smallskip}
			& \multicolumn{3}{c}{{\bf U4 dataset}} & \multicolumn{3}{c}{{\bf Stanford Drone dataset}} \\
			\noalign{\smallskip}
			\hline
			\noalign{\smallskip} 
			& {\bf Training} & & {\bf Inference} & {\bf Training} & & {\bf Inference}  \\
			\noalign{\smallskip}
			\parbox[t]{2mm}{\multirow{3}{*}{\rotatebox[origin=c]{90}{IOCMM}}} & Traj. batch $B_t$: 10 & Stoch. policy $\alpha$: 0.1 & $s_{0,g}$ sampl. $\tau$: 0.01 & Traj. batch $B_t$: 50 & Stoch. policy $\alpha$: 0.1 & $s_{0,g}$ sampl. $\tau$: 0.01 \\
			& Map batch $B_m$: 7 & Learning rate $\lambda$: 1.0  & Num. sim. traj.: 500 & Map batch $B_m$: 5 & Learning rate $\lambda$: 1.0 & Num. sim. traj.: 1000 \\
			& Base reward $r_0$: 0.01 & & & Base reward $r_0$: 0.01 & & \\
			\noalign{\smallskip}
			\hline
			\noalign{\smallskip}
			\parbox[t]{2mm}{\multirow{7}{*}{\rotatebox[origin=c]{90}{Semapp}}} & Pooling layers: 1 & Crop size: 32 & Num. crops: 1 & Pooling layers: 1 & Crop size: 64 & Num. crops: 500 \\
			& Growth rate: 2 & Batch size: 32 & & Growth rate: 2 & Batch size: 32 & \\
			& Layers per block: 5 & Dropout prob.: 0.35 & & Layers per block: 5 & Dropout prob.: 0.35 & \\
			& Num. conv. layers: 19 & Learning rate: 0.01 & & Num. conv. layers: 19 & Learning rate: 0.01 & \\
			& Num. param.: 6.5 k & Learn. rate decay: 0.9985 & & Num. param.: 6.5 k & Learn. rate decay: 0.9985 & \\
			& \multicolumn{2}{l}{Weight decay: $4.5 \times10^{-5}$} & & \multicolumn{2}{l}{Weight decay: $4.5 \times10^{-5}$} & \\
			\noalign{\smallskip}
			\hline
		\end{tabular}
	\end{center}
\end{table*}
\setlength{\tabcolsep}{1.4pt}

\begin{figure}[t]
	\begin{center}
		\vspace{5pt}
		\includegraphics[angle=270,width=0.32\columnwidth]{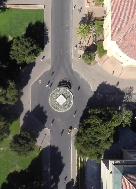}
		\includegraphics[angle=270,width=0.32\columnwidth]{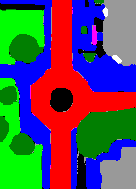}
		\includegraphics[angle=270,width=0.32\columnwidth]{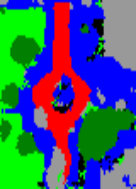} \\
		\vspace{3pt}
		\includegraphics[angle=270,width=0.32\columnwidth]{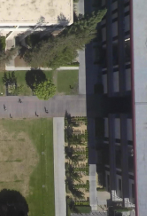}
		\includegraphics[angle=270,width=0.32\columnwidth]{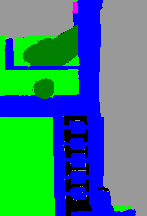}
		\includegraphics[angle=270,width=0.32\columnwidth]{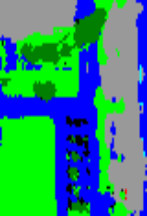} \\
	\end{center}
	\vspace{-5pt}
	\caption{Semantic segmentation, obtained by our proof-of-concept pipeline in the Stanford Drone dataset. {\bf Left:} input images. {\bf Middle:} ground truth labels. {\bf Right:} predicted semantic map.}
	\label{fig:segmentation}
	\vspace{-8pt}
\end{figure}

The \emph{Stanford Drone Dataset} (SDD) \cite{robicquet2016learning} was recorded on the Stanford University grounds, which include a wide variety of environments and semantic classes, e.g. shared roads for cyclists and vehicles, pedestrian areas, college buildings, vegetation and parking lots. The dataset includes 51 top-down scenes with bounding boxes for various agents, from which we extracted trajectories of people, approximating the position by the center of the bounding box. We chose 25 scenes sufficiently covered by trajectories and scaled the maps to the constant physical resolution of 0.4 m per cell. We manually segmented each scene into nine semantic classes: pedestrian area, vehicle road, bicycle road, grass, tree foliage, bulging, entrance, obstacle and parking. Some example scenes from SDD are shown in Fig.~\ref{fig:cover}, \ref{fig:large_maps} and \ref{fig:segmentation}.

\subsection{Training and Evaluation}
\label{sec:experiments:training}

To our knowledge, there exist quite some works on human motion prediction \cite{rudenko2020surveyIJRR} but none of the existing methods predicts prior occupancy distribution of walking people in urban environments only based on semantic information -- a task considerably different from trajectory prediction \cite{kitani2012activity,karasev2016intent}. Therefore in our experiments we mainly compare the IOC and CNN solutions to the problem against the ground truth distributions and the following baseline methods:
\begin{enumerate}
	\item uniform distribution over $\mathcal{M}$
	\item uniform distribution over the walkable states in $\mathcal{M}$
	\item semantics-unaware \emph{mapp} network \cite{doellinger2018predicting}
\end{enumerate}
For quantitative evaluation we measure \emph{Kullback-Leibler divergence} (KL-div) between the predicted and the ground truth distribution:
\begin{equation}
	D_{\text{KL}}(P_{\text{GT}}||Q_{\text{Pred.}}) = \sum_{x \in \mathcal{M}} P_{\text{GT}}(x) \log  \frac{P_{\text{GT}}(x)}{Q_{\text{Pred.}}(x)}.
	\label{eq:kl-div}
\end{equation}

We train and evaluate IOCMM, semapp and the baselines separately on the U4 and Stanford Drone datasets. Training and inference parameters are summarized in Table~\ref{tab:hyperparamaters}. 

We optimized the hyperparameters for IOCMM prior to the main experiments on a small portion of data from both datasets. For KL-div benchmarking we learn the $\bm{\theta}$ weights from leave-one-out maps in the dataset, and validate the result in the remaining map, iterating over all maps in the respective dataset. Furthermore, we evaluate the impact of the number of simulated trajectories for inference in the new map (as in Eq.~\ref{eq:occupancy_estimation}), measuring runtime and solution quality. Finally, we estimate the two $s_{0,g}$ sampling strategies, detailed in Sec.~\ref{sec:ioc}, separately.

Training the mapp and semapp networks on the U4 dataset is straightforward as the size of the maps (32 by 32 pixels) is equivalent to the network input. From the larger images in the SDD (on average 146 by 152 pixels) we take 500 random crops of size 64 by 64 pixels. Each crop in the training data is augmented 7 times by rotation and mirroring. In both cases, leave-one-out maps are used for training and validation (50/50), and the remaining map is used for evaluation.
We followed the training procedure and hyperparameters from \cite{doellinger2018predicting} which turned out robust enough for our application. In particular, we trained the networks for 100 epochs with binary cross-entropy loss using the Adam optimizer \cite{kingma2014adam}, and stopped the training when the performance on the validation set did not improve for 15 epochs. Further parameters are detailed in Table~\ref{tab:hyperparamaters}.

\subsection{Semantic Segmentation}
\label{sec:experiments:segmentation}

As a proof-of-concept that the semantic maps, required for the methods presented in this paper, can be obtained from images during runtime, we have set up a preliminary pipeline for semantic segmentation using UNet implementation in Keras and TensorFlow \cite{semanticSegmentationKeras}. We trained the CNN with the 19 images from the Stanford Drone dataset, augmented 3 times with rotation, and tested on the remaining 6 images. Even with such negligible amount of data, our experiments (see Fig.~\ref{fig:segmentation}) reached 0.64 frequency-weighted IoU in the training dataset (78\% accuracy), and 0.53 IoU (69\% accuracy) in the test dataset. We are confident, that using state-of-the-art semantic segmentation techniques \cite{cityscapesLeaderboard} the performance, necessary for the application of our method, will be reached. Combining these two pipelines is of prime priority for our future work.

\section{Results}
\label{sec:results}

\begin{figure*}[t]
	\begin{center}
		\vspace{5pt}
		\includegraphics[width=1.55\columnwidth]{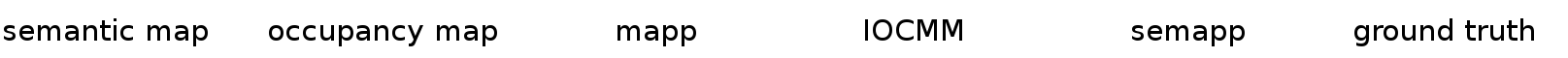} \\
		\includegraphics[width=0.25\columnwidth]{media/toy/outfile_result_toy_14_crop_1_sem.png}
		\includegraphics[width=0.25\columnwidth]{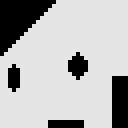}
		\includegraphics[width=0.25\columnwidth]{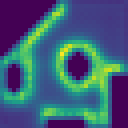}
		\includegraphics[width=0.25\columnwidth]{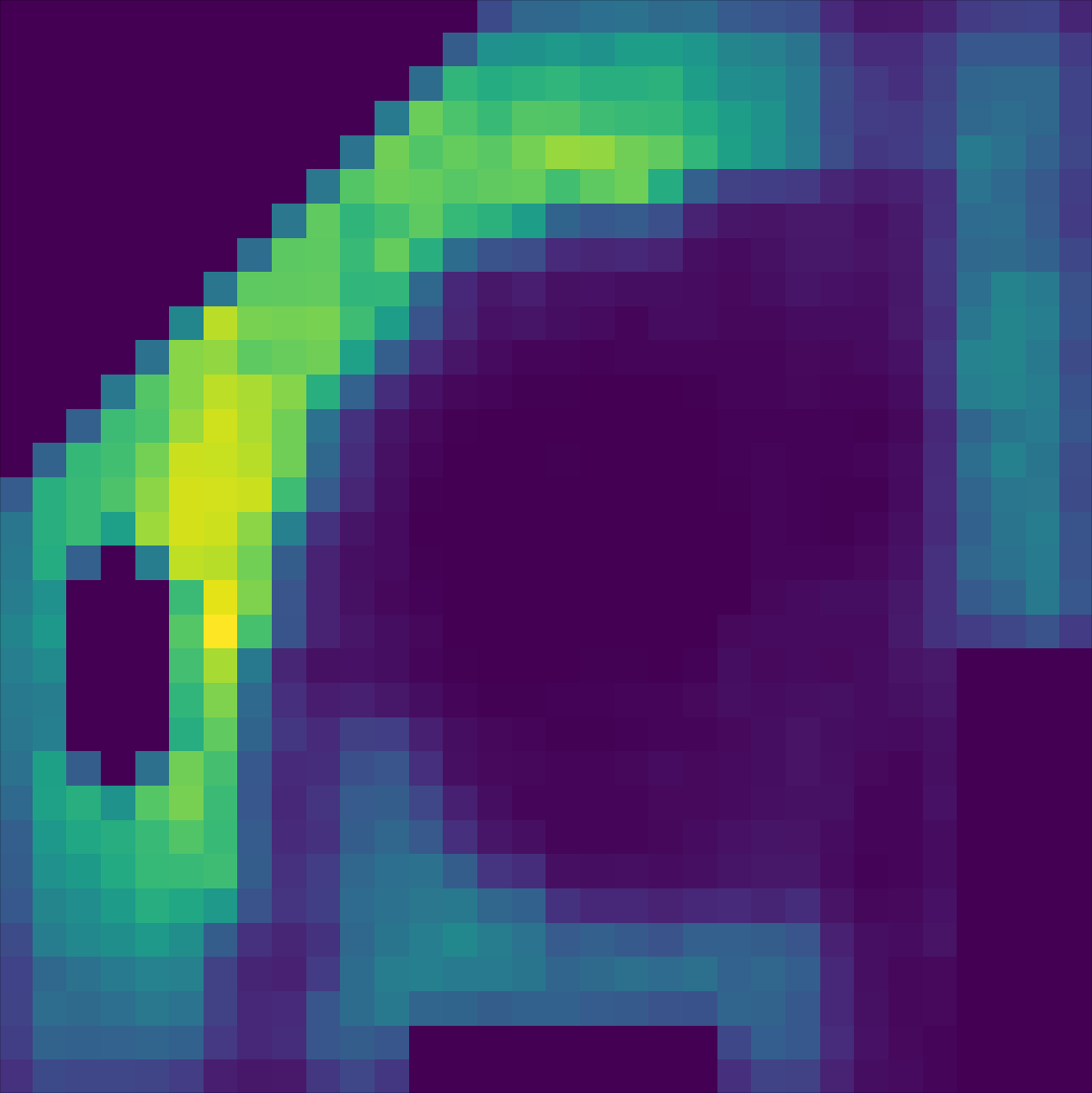}
		\includegraphics[width=0.25\columnwidth]{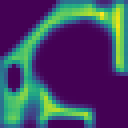}
		\includegraphics[width=0.25\columnwidth]{media/toy/outfile_result_toy_14_crop_1_gt.png} \\
		\vspace{2pt}
		\includegraphics[width=0.25\columnwidth]{media/toy/outfile_result_toy_33_crop_1_sem.png}
		\includegraphics[width=0.25\columnwidth]{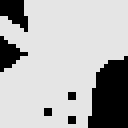}
		\includegraphics[width=0.25\columnwidth]{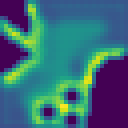}
		\includegraphics[width=0.25\columnwidth]{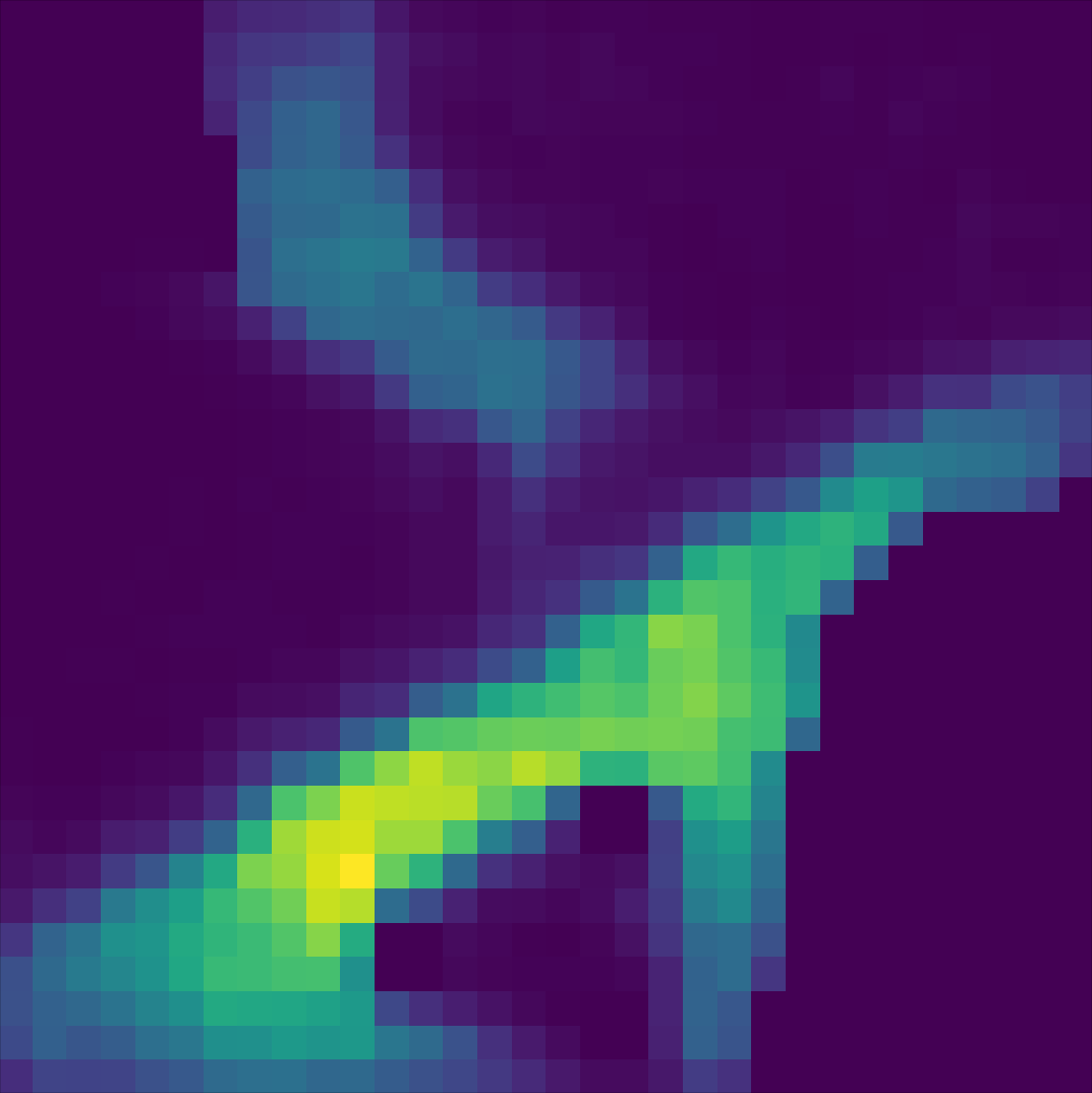}
		\includegraphics[width=0.25\columnwidth]{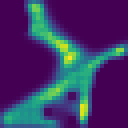}
		\includegraphics[width=0.25\columnwidth]{media/toy/outfile_result_toy_33_crop_1_gt.png} \\
		\vspace{2pt}
		\includegraphics[width=0.25\columnwidth]{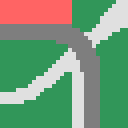}
		\includegraphics[width=0.25\columnwidth]{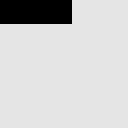}
		\includegraphics[width=0.25\columnwidth]{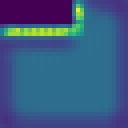}
		\includegraphics[width=0.25\columnwidth]{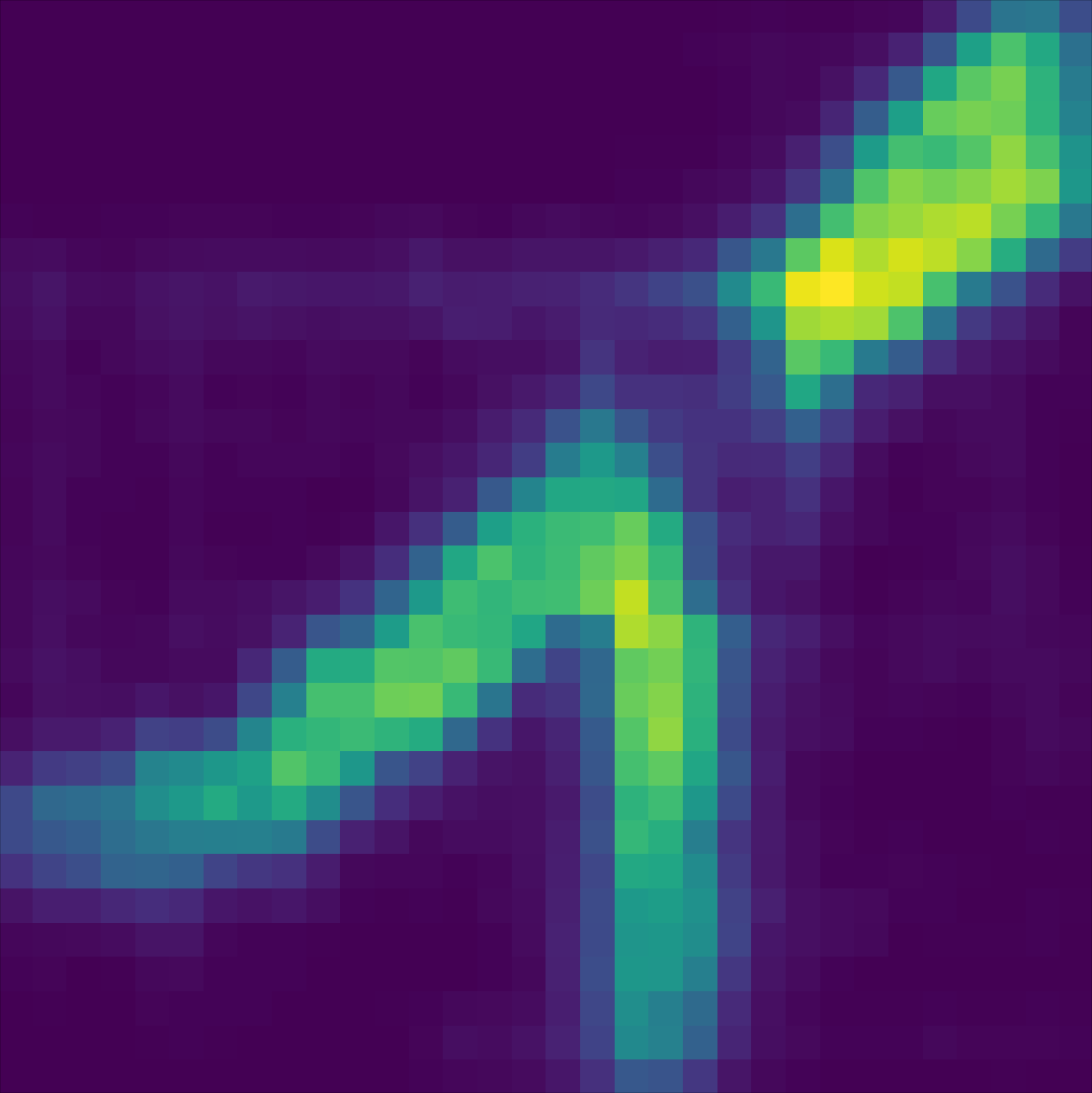}
		\includegraphics[width=0.25\columnwidth]{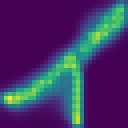}
		\includegraphics[width=0.25\columnwidth]{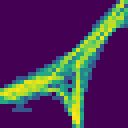} \\
		\vspace{2pt}
		\includegraphics[width=0.25\columnwidth]{media/toy/outfile_result_toy_2_crop_1_sem.png}
		\includegraphics[width=0.25\columnwidth]{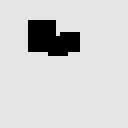}
		\includegraphics[width=0.25\columnwidth]{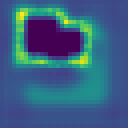}
		\includegraphics[width=0.25\columnwidth]{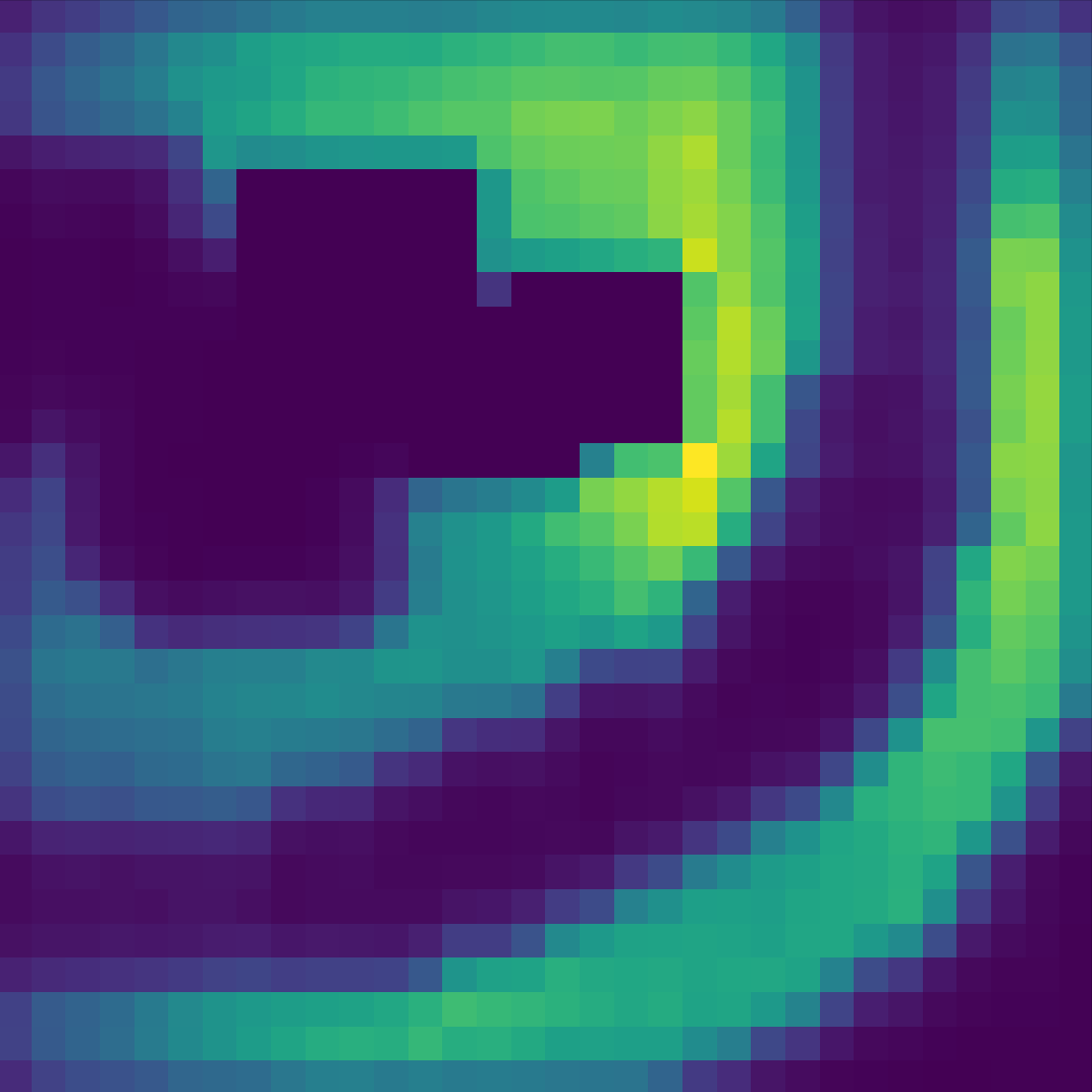}
		\includegraphics[width=0.25\columnwidth]{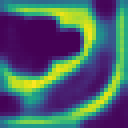}
		\includegraphics[width=0.25\columnwidth]{media/toy/outfile_result_toy_2_crop_1_gt.png} 
	\end{center}
	\vspace{-8pt}
	\caption{Qualitative comparison of results in the U4 dataset. A binary \emph{occupancy map} of the environment highlights the amount of structure imposed by semantics in urban scenes. There is little surprise that the semantics-unaware mapp approach for learning occupancy priors \cite{doellinger2018predicting} is not learning any meaningful behaviors apart from the fact that people (often) tend to be found close to obstacles. On the contrary, IOCMM correctly estimates the priors in different walkable areas. On top of that, CNN-based semapp is capable of detecting all ``illegal crosswalks'' in these scenes, as well as cutting over grass in such places where the topology of the environment encourages to do so, e.g. see the sharp corner in the third row.}
	\label{fig:results_comparison}
	\vspace{-8pt}
\end{figure*}

We report the mean and standard deviations of the KL-divergences for both datasets in Table~\ref{tab:results}. In the U4 dataset, both IOCMM and semapp outperform the other baselines, furthermore both proposed sampling strategies for IOCMM show similar performance after appropriate hyperparameter optimization. Semapp, in addition to the quantitative improvement of minimum 14\% over the closest baseline (IOCMM), offers a clear qualitative improvement in identifying crucial non-linearities in the predicted priors, as displayed in Fig.~\ref{fig:results_comparison}. This figure shows the extent to which semantics of the environment impact the distribution prediction -- all semantics-unaware methods in our comparison, e.g. uniform $p(s)$ over $\mathcal{M}$ and mapp, perform poorly. On the contrary, in both datasets semapp outperforms all baselines, due to its ability to reason over spatially-connected regions using convolutions, learning not only local contexts where motion probability is high, but also which locations are usually avoided by pedestrians. Interestingly, in the SDD dataset, due to incomplete ground truth coverage of the scenes (as seen in Fig.~\ref{fig:cover} and \ref{fig:large_maps}), removing unwalkable spaces from the uniform distribution over all states in $\mathcal{M}$ only decreases performance of this baseline. The reason here is that for many walkable states, where no motion is recorded in the ground truth, probabilities increase, resulting in worse KL-div scores. Despite using this imperfect training material, semapp consistently outperforms all baselines with the best KL-div score and smaller standard deviation between maps.

\setlength{\tabcolsep}{4pt}
\begin{table}
	\begin{center}
		\caption{Average KL-div in the U4 and Stanford Drone datasets}
		\label{tab:results}
		\scriptsize
		\begin{tabular}{lll}
			\hline\noalign{\smallskip}
			Method & U4 dataset & Stanford Drone \\
			\noalign{\smallskip}
			\hline
			\noalign{\smallskip} 
			Uniform $p(s)$ over $\mathcal{M}$ & $1.21 \pm 0.26$ & $1.40 \pm 0.31$ \\
			Uniform $p(s)$ over walkable states in $\mathcal{M}$ & $0.97 \pm 0.28$ & $1.69 \pm 0.42$ \\
			Uniform $p(s|\bm{f}(s))$ learned from $\bm{\mathcal{T}}_\mathcal{M}$ & $0.53 \pm 0.09$ & $1.09 \pm 0.26$ \\
			mapp CNN & $ 0.93 \pm 0.27 $ & $ 1.03 \pm 0.19 $ \\
			IOCMM with learned $p(s_{0,g})$ & $0.42 \pm 0.07$ & $ 1.04 \pm 0.24 $  \\
			IOCMM with modeled $p(s_{0,g})$ & $0.43 \pm 0.07$ & $ 1.21 \pm 0.32 $  \\
			\noalign{\smallskip}
			semapp CNN & $\bm{0.37 \pm 0.11}$ & $\bm{0.71 \pm 0.13}$ \\ \hline
		\end{tabular}
	\end{center}
\end{table}
\setlength{\tabcolsep}{1.4pt}

\begin{figure}[t]
	\begin{center}
		\vspace{5pt}
		\includegraphics[width=0.423\columnwidth]{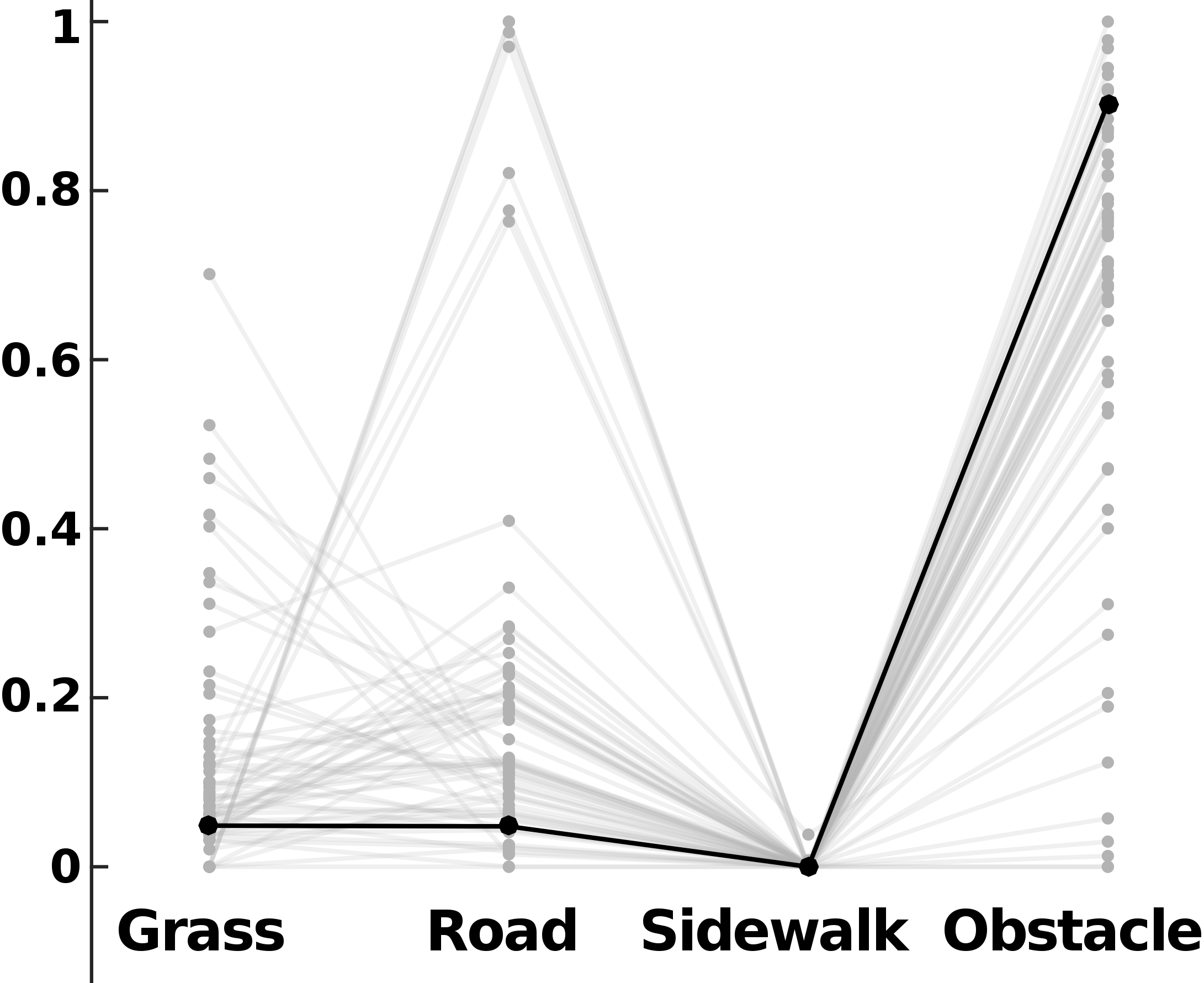}
		\includegraphics[width=0.583\columnwidth]{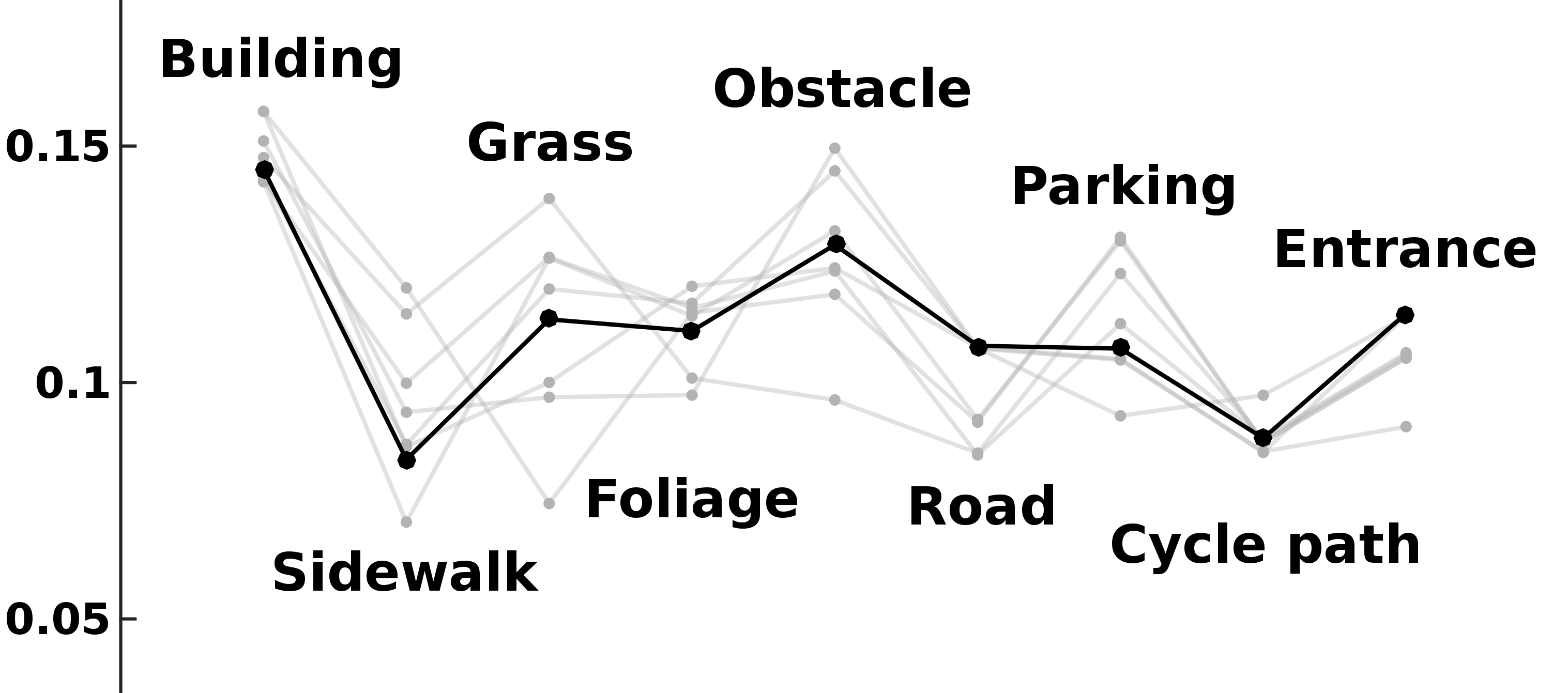}
	\end{center}
	\vspace{-8pt}
	\caption{Optimal $\bm{\theta}$ costs of various semantic classes, learned by IOCMM in each individual map, are shown in gray. Globally optimal weights for the entire dataset are overlaid in black. {\bf Left:} U4 dataset. {\bf Right:} Stanford Drone dataset.}
	\label{fig:thetas_toy}
	\vspace{-8pt}
\end{figure}

In Fig.~\ref{fig:thetas_toy}, we visualize the generalization capabilities of the IOCMM method. To this end, we show the optimal $\bm{\theta^*}$ weights in each individual map in both datasets, and compare them to the globally optimal set of weights, learned from training on all maps in the respective dataset. Here lies one benefit of the Inverse Optimal Control strategy to find occupancy priors: IOCMM does not only generalize well on a large amount of maps, but also retains high performance when learning from small amounts of data. In fact, when training on a fraction of maps from the dataset (e.g. as little as 10 random maps), IOCMM on average still converges to the globally-optimal $\bm{\theta}^*$ costs for semantic classes, and thus the KL-div scores do not drop. This property is not shared by semapp, which needs a large selection of maps sufficiently covered by trajectories to generalize across various local contexts.

\begin{figure}[t]
	\begin{center}
		\vspace{5pt}
		\hspace{13pt}
		\includegraphics[width=0.493\columnwidth]{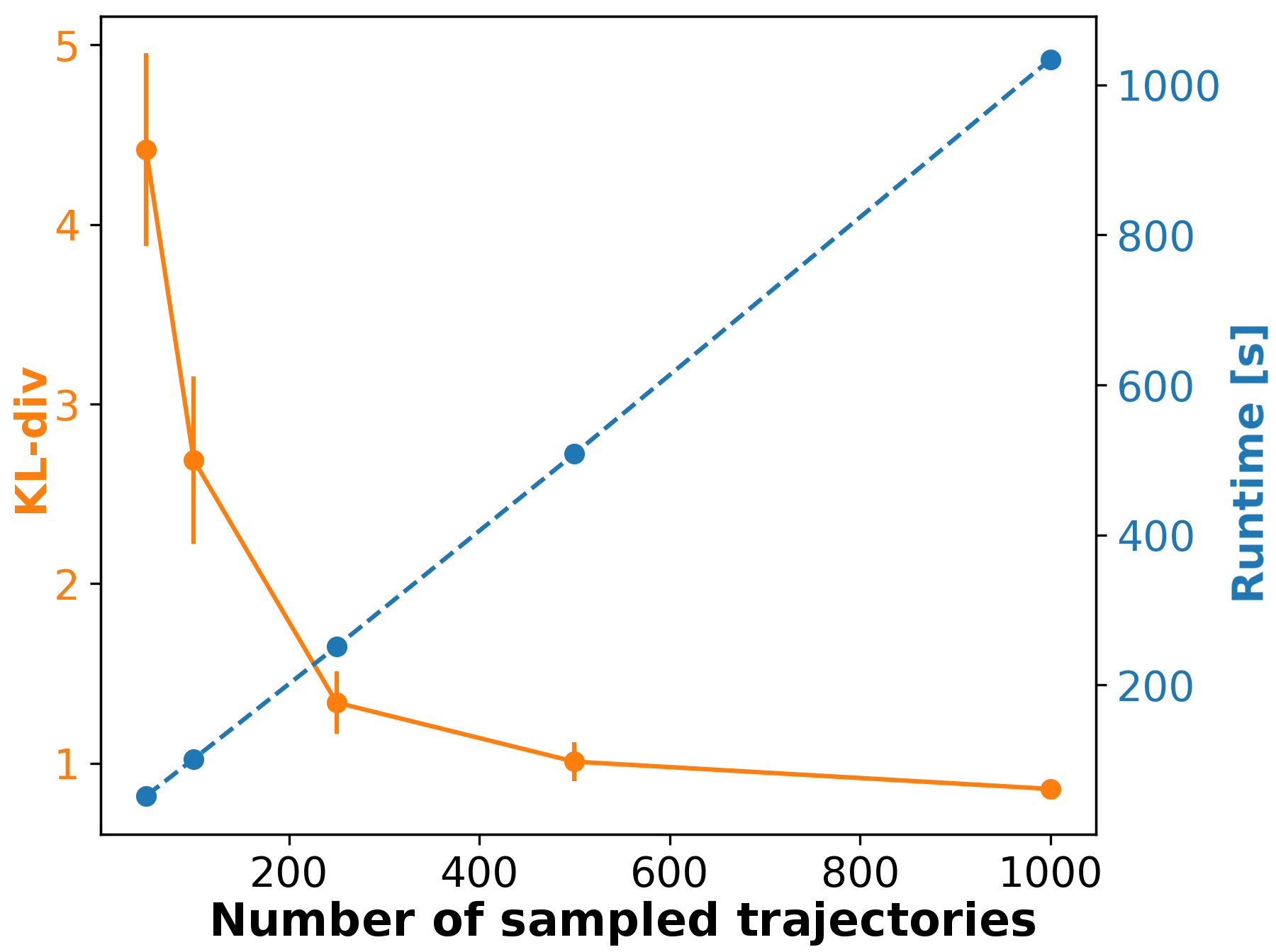}
		\includegraphics[width=0.493\columnwidth]{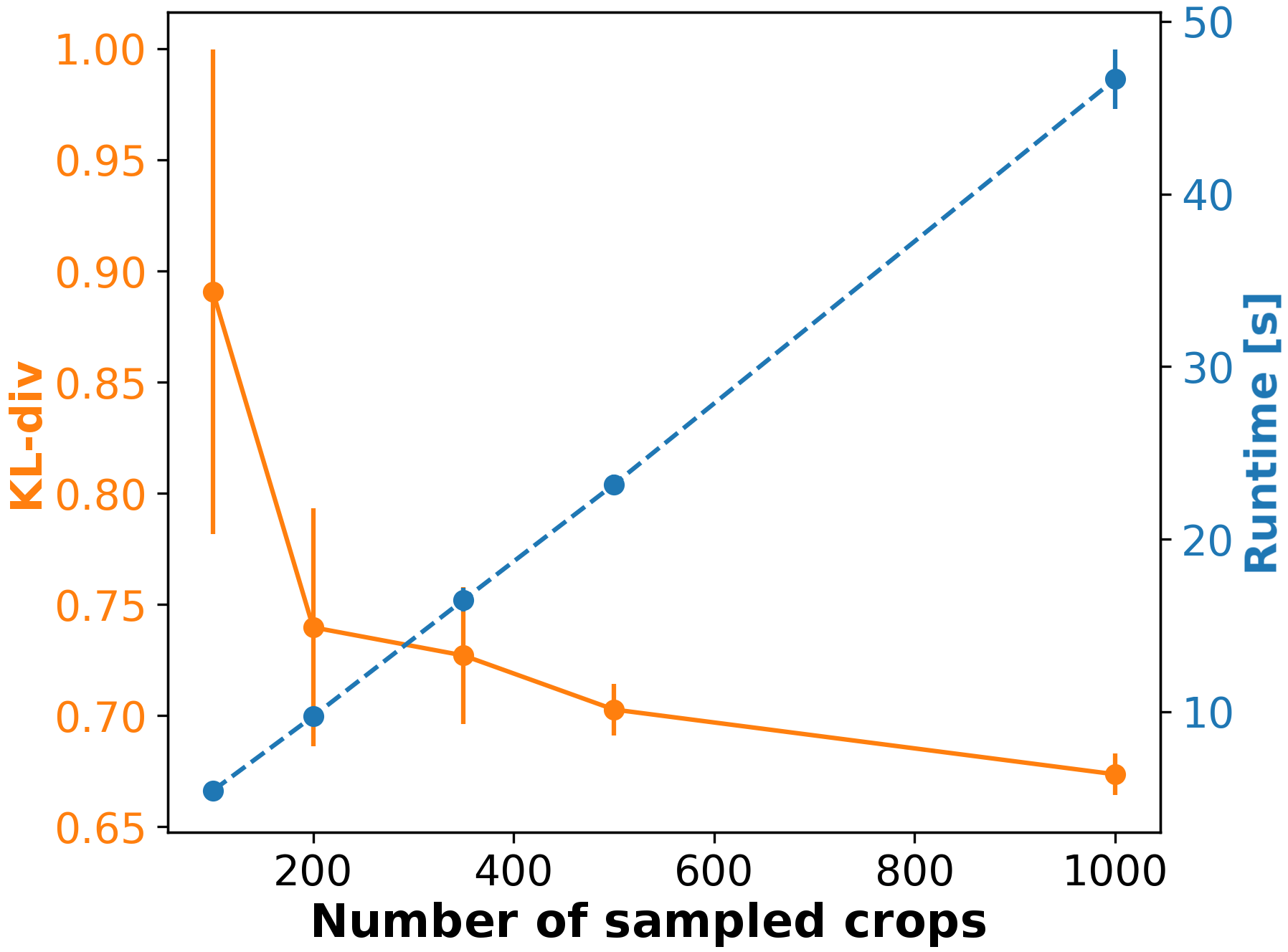}
	\end{center}
	\vspace{-8pt}
	\caption{Runtime and performance of IOCMM {\bf (left)} and semapp {\bf (right)} in the SDD dataset as a function of the number of sampled trajectories/crops during inference. Both methods' runtime scales linearly with the number or random samples, while performance improves exponentially.}
	\label{fig:kl_runtime}
	\vspace{-8pt}
\end{figure}

Both methods, IOCMM and semapp, depend on the number of random samples during inference. IOCMM samples trajectories between random start and goal positions, while semapp samples random crops from the larger semantic map. In Fig.~\ref{fig:kl_runtime} we show the relation between the number of samples, performance and inference time in the large maps of the SDD dataset. Runtimes were measured for Python implementations of both algorithms on an ordinary laptop with Intel Xeon 2.80GHz $\times~8$ CPU and 32 GB of RAM. The CNN is implemented using Theano on the built-in GPU Quadro M2000M. It is worth mentioning that the inference time of semapp on one crop of size 64 by 64 pixels (equivalent to $25.6 \times 25.6$ m) is $\sim 0.054$ seconds, appropriate for real-time application.

\section{Conclusion}
\label{sec:discussion}

In this paper, we research the problem of learning human occupancy priors in semantically-rich urban environments using only the semantic map as input. Considering two established classes of approaches to this end (Inverse Optimal Control and Convolutional Neural Networks), we show that our CNN-based semapp approach is outperforming all baselines already with limited training data.
The IOCMM approach, on the other hand, can be used to reasonably estimate the costs of semantic classes from several maps and few trajectories. However, it is limited to constant weights, which may not reflect behavior of people in all local contexts of semantically-complex environments. This approach lacks reasoning on spatial relevance of surfaces to infer cases where people may prefer to walk on one surface class over another, or not walk at all.

In future work we intend to further investigate the possibilities of applying advanced IOC techniques, for instance non-linear IRL with complex features \cite{levine2011nonlinear}, non-linear reward modeling \cite{mainprice2016functional}, automated feature extraction to exploit local correlations in the environment \cite{levine2010feature,metelli2017compatible}, IOC with multiple locally-consistent reward functions \cite{nguyen2015inverse} and with CNN-based reward function approximator \cite{wulfmeier2015maximum}. Furthermore, we plan to validate semapp with on-the-fly semantics estimation and extend it to first-person view for application in automated driving to infer potential pedestrians' entrance points to the road.




\bibliographystyle{IEEEtran}
\footnotesize{
	\bibliography{bibliography}
}

\end{document}